\documentclass{article}

\usepackage{arxiv}

\usepackage[utf8]{inputenc} % allow utf-8 input
\usepackage[T1]{fontenc}    % use 8-bit T1 fonts
\usepackage{hyperref}       % hyperlinks
\usepackage{url}            % simple URL typesetting
\usepackage{booktabs}       % professional-quality tables
\usepackage{amsfonts}       % blackboard math symbols
\usepackage{nicefrac}       % compact symbols for 1/2, etc.
\usepackage{microtype}      % microtypography
\usepackage{lipsum}
\usepackage{graphicx}
\usepackage{xcolor}
\usepackage{amsmath}
\graphicspath{ {./images/} }

\begin{document}

%%%%%%%%% TITLE - PLEASE UPDATE
\title{Adaptive Transfer Learning: a simple but effective transfer learning}

\author{
  Jung H Lee \\
  Pacific Northwest National Laboratory,\\
  Seattle, WA USA \\
  {\tt\small jung.lee@pnnl.gov}

  \and 
  Henry J Kvinge  \\
  Pacific Northwest National Laboratory,\\
  Seattle, WA USA \\
  {\tt\small henry.kvinge@pnnl.gov}
  
  \and 
  Scott Howland \\
  Pacific Northwest National Laboratory,\\
  Seattle, WA USA \\
  {\tt\small scott.howland@pnnl.gov}

  \and
  Zachary New\\
  Pacific Northwest National Laboratory,\\
  Seattle, WA USA \\
  {\tt\small zachary.new@pnnl.gov}
  
  \and
  John Buckheit\\
  Pacific Northwest National Laboratory,\\
  Seattle, WA USA \\
  {\tt\small john.buckheit@pnnl.gov} 
  
  \and
  Lauren A. Phillips\\
  Pacific Northwest National Laboratory,\\
  Richland, WA USA \\
  {\tt\small lauren.phillips@pnnl.gov}
 
  \and
  Elliott Skomski\\
  Pacific Northwest National Laboratory,\\
  Seattle, WA USA \\
  {\tt\small elliott.skomski@pnnl.gov}
  
  \and
  Jessica Hibler\\
  Pacific Northwest National Laboratory,\\ 
  Richland, WA USA\\
  {\tt\small jessica.hibler@pnnl.gov}
  
  \and
  Courtney D. Corley\\
  Pacific Northwest National Laboratory,\\ 
  Richland, WA USA\\
  {\tt\small court@pnnl.gov}

  \and
  Nathan O. Hodas\\
  Pacific Northwest National Laboratory,\\ 
  Richland, WA USA\\
  {\tt\small nathan.hodas@pnnl.gov}
}

\maketitle

%%%%%%%%% ABSTRACT
\begin{abstract}
   Transfer learning (TL) leverages previously obtained knowledge to learn new tasks efficiently and has been used to train deep learning (DL) models with limited amount of data. When TL is applied to DL, pretrained (teacher) models are fine-tuned to build domain specific (student) models. This fine-tuning relies on the fact that DL model can be decomposed to classifiers and feature extractors, and a line of studies showed that the same feature extractors can be used to train classifiers on multiple tasks. Furthermore, recent studies proposed multiple algorithms that can fine-tune teacher models’ feature extractors to train student models more efficiently. We note that regardless of the fine-tuning of feature extractors, the classifiers of student models are trained with final outputs of feature extractors (i.e., the outputs of penultimate layers). However, a recent study suggested that feature maps in ResNets across layers could be functionally equivalent, raising the possibility that feature maps inside the feature extractors can also  be used to train student models’ classifiers. Inspired by this study, we tested if feature maps in the hidden layers of the teacher models can be used to improve the student models’ accuracy (i.e., TL’s efficiency). Specifically, we developed `adaptive transfer learning (ATL)’, which can choose an optimal set of feature maps for TL, and tested it in the few-shot learning setting.  Our empirical evaluations suggest that ATL can help DL models learn more efficiently, especially when available examples are limited.
\end{abstract}

%%%%%%%%% BODY TEXT
\section{Introduction}

Deep learning (DL) has been used in a broad range of domains, as it learns abstract rules from examples \cite{lecun_deep_2015, lecun_gradient-based_1998, hertz_introduction_nodate}. For instance, DL can generalize visual features from given examples, which are superior to those selected by human experts. However, DL requires a huge amount of labelled examples and computing resources to learn complex tasks, making its deployment in the real world challenging. Transfer learning (TL), which leverages previously obtained knowledge from one task to learn new tasks more efficiently, may alleviate the demanding requirements of DL \cite{pan_survey_2010, yosinski_how_nodate, azizpour_generic_nodate, kornblith_better_nodate, tajbakhsh_convolutional_2016, razavian_cnn_nodate}. Specifically, when TL is applied to DL, the features/feature maps of pretrained \textit{teacher} models are reused to build new \textit{student} models. 

Interestingly, it has been thought that initial layers of deep neural networks (DNNs), closer to input layers, learn low-level features (e.g., shapes and edge), and top layers, closer to output layers, learn high-level complex features (e.g., eye and ears) \cite{yosinski_how_nodate}. Further, the high-level visual features can often be shared between tasks. For example, cats’ eyes can be used to detect dogs’ eyes. Inspired by this line of thoughts, the last layer (or a few of top layers) of teacher models, which are pretrained on `source’ tasks, are fine-tuned to  build student models on new `target’ tasks. In this fine-tuning (FT) of teacher models, the outputs of feature maps in penultimate layers are used to train final classification layers (FCL) on target tasks. A line of studies suggests that FT, a realization of TL in DL, can train student models rapidly with limited amount of examples, raising the possibility that FT could effectively address the high cost of DL; see \cite{pan_survey_2010} for a review. 

However, it is not guaranteed that high-level features in the top layers (e.g., penultimate layers) are the optimal features to learn new target tasks. If target and source tasks are significantly different, low-level features would be more suitable to reuse than the high-level ones. Further, an earlier study \cite{veit_residual_nodate} suggested that ResNet behaves like a collection of shallow networks, which indicates that feature maps in the initial layers can be as valuable as those in top layers. Then, the question is: how do we select the best features? We hypothesized that the best feature maps for target tasks would selectively respond to the classes of target tasks. For instance, if student models need to learn about canine breeds, teacher models’ feature maps, which respond to retrievers but not to corgis, would be the best ones to transfer to student models. To address this hypothesis, we developed `adaptive transfer learning (ATL)’ that uses response characteristics of feature maps to identify the best features for target tasks. Once the best feature maps are selected, their outputs are used to train a final classification layer (FCL). In this study, we used ResNet50 pretrained on ImageNet as teacher models, since ResNet \cite{he_deep_2015} may contains equally valuable feature maps across layers \cite{veit_residual_nodate}. 

Since TL is commonly used to train DL models with limited amount of training examples, we focused on evaluating ATL's advantage in the few-shot learning. To this end, we synthesized the few-shot learning problems by selecting a subset of classes in 5 popular benchmark problems (CUB\_200\_2011 \cite{WahCUB_200_2011}, FGVC-Aircraft \cite{maji13fine-grained}, CIFAR100 \cite{krizhevsky_learning_nodate}, Fruits\_360 \cite{article} and Omniglot \cite{Lake1332}). Specifically, we randomly drew 30 classes from the benchmark datasets and used 5, 10, 15, 20, 25 and 30 classes to mimic 5, 10, 15, 20, 25 and 30-way learning. After choosing the classes, 3, 5 and 10 examples were randomly chosen from the training set as training examples to mimic 3, 5 and 10-shot learning. To evaluate the benefit of ATL, we measured the accuracy of FCL on all available test examples, and our empirical evaluations suggest that ATL can outperform the baseline model, in which the last layer is fine-tuned on target tasks. Based on our results, we propose that ATL can be a simple yet effective solution to DL's demanding requirements of labelled data.

\section{ATL and its evaluation}

\subsection{Related Works}

FT falls into three categories depending on the training methods of student models. First, only the last synaptic layer from penultimate to classification layers is fine-tuned on target tasks \cite{simonyan_very_2015}. In this approach, the feature extractor is not fine-tuned. Second, a few top layers of the student models are fine-tuned on target tasks. As a few layers of feature extractors are fine-tuned, the feature maps used to train student models’ classifiers are also customized for target tasks\cite{pmlr-v32-donahue14, 10.1007/978-3-319-10590-1_53}. Third, all layers of student models can be retrained. Although all layers are retrained, pretrained weights of teacher models allow student models to learn fast with limited amount of examples. 

Notably, regardless of  the number of layers retrained, the outputs of penultimate layers are used to train FCL on target tasks. The outputs of penultimate layers (i.e., the feature maps in the layer) are flattened to train fully connected classifiers. Multiple studies explored the most effective way to utilize penultimate layers' features. \cite{8099808, song_locally-transferred_2017, gao_compact_2016, cimpoi_deep_2015}. For instance, high-order interactions between features were estimated by bilinear pooling \cite{7410527} to train FCL more accurately. Also, two recent studies proposed the algorithms that can dynamically freeze or unfreeze layers (Adaptive filter \cite{guo_adafilter_2019}) or feature maps (Spot Tune \cite{guo_spottune_nodate}) in teacher models. Despite differences in details, both algorithms trained the secondary network models to determine the best policy of feature map selection and thus require sufficient amount of data to train the secondary networks. 

\begin{figure*}[t]
\centering
\includegraphics[width=1\textwidth]{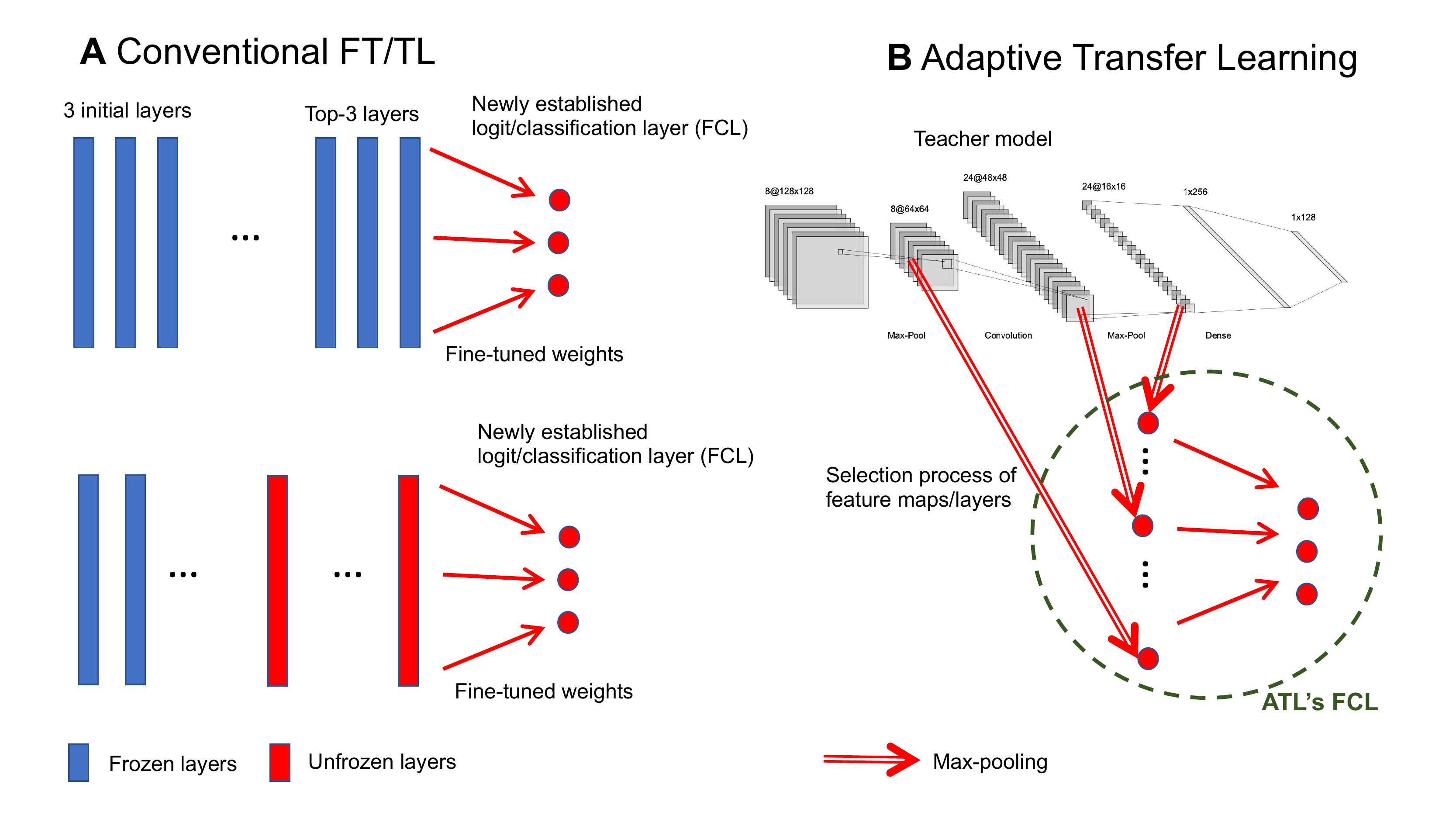}
\caption{Schematics of TL in (A) and adaptive transfer learning in (B). Teacher models’ final classification layers are rebuilt to perform target tasks. In the conventional TL, a subset of layers are unfrozen and retrained. If only the final (synaptic) layer is retrained, the same feature maps obtained from the source tasks are used for the newly established final classification layer (FCL). If multiple layers are retrained, feature maps in the penultimate layer are fine-tuned for FCL. That is, only the feature maps in the penultimate layers are used for FCL. In contrast, the adaptive transfer learning can use feature maps across layers for FCL. The max-pooling is used to map each feature onto a single input node for FCL.}
\label{fig1} 

\end{figure*}

We note that ResNet behaves like a collection of shallow networks \cite{veit_residual_nodate} and that earlier layers (than penultimate layers) of teacher models were found to be more valuable than the late layers for digital pathology (biomedical research) \cite{8575475}. Inspired by these studies, we sought algorithms to select the optimal features/feature maps across all layers of teacher models. Fig. \ref{fig1} illustrates our newly proposed approach (ATL). ATL first evaluates the relevance of individual layers to target tasks and select the best (most selective) feature maps in relevant layers. Finally, the outputs of best feature maps are used to train FCL on target tasks. In this study, ResNet50, pretrained on ImageNet, is used as a teacher model. Below, we describe our selection rules of relevant layers and feature maps in detail and discuss the experimental protocols. 

\subsection{Evaluation of relevance of layers to the target tasks}
Inspired by earlier studies \cite{snell_prototypical_2017,abdelkader_headless_2020} suggesting that penultimate layers’ activations can reflect the inputs’ classes (i.e., labels), we hypothesized that the relevance of layers to the target tasks could be estimated by examining how well the layers’ outputs are clustered according to the inputs’ classes. That is, if the outputs are well clustered, the selected layers are relevant. In the experiments, we measured how well the layer’s outputs are clustered in three steps. First, we estimated feature maps’ activations induced by training examples and determined the maximal activations in individual maps (Fig. \ref{fig1}B). These maximal activations (i.e., max-pooling) were used to represent the feature maps’ outputs, and we constructed the layers’ reduced activation vector ($LAV^r $) by aggregating the feature maps’ maximal activations. Second, we estimated the centroids of classes by using normalized $LAV^r$ (Eq. \ref{eq1}). 

\begin{equation}\label{eq1}
C_{k}=\frac{1}{N}\Sigma_{i=1,N}\frac{LAV^r_i}{\left\Vert LAV^r_i  \right\Vert},
\end{equation}
where $N$ and $k$ represent the number of examples and the selected class. 

Third, we estimated the distances between all pairs of classes' centroids to obtain the minimum distances (Eq. \ref{eq2}) in layers, which indicates the layers’ relevance scores ($R$); higher minimum distances indicate more relevant layers. We used R to select the most relevant layers. The exact number of layers is a hyperparameter. In the experiments, top-3 relevant layers were used unless stated otherwise. 

\begin{equation}\label{eq2}
R=min(\left\Vert C_i, C_j \right \Vert),
\end{equation}

where $C_{i,j}$ represents the centroid of the class i, j (Eq. \ref{eq1}).

\subsection{Selection of feature maps in relevant layers}
After selecting the relevant layers, we tested if the feature maps in the layers selectively respond to any class of training examples by using `t-test’. If feature maps selectively respond to any class, their responses significantly vary depending on the inputs’ classes. For instance, if a feature map is selective to class 1, its responses to class 1 examples will differ from those to other class examples. The t-test computes the probability ($p$-value) that the two distributions are drawn from the same distribution. That is, lower $p$-values indicate that the two distributions are significantly different. To use this t-test to evaluate feature map’s selectivity, we estimated $LAV^r$ to all training examples and used them to build two distributions; that is, we considered the maximal outputs of the feature maps. The first distribution contains $LAV^r$ induced by a target class, whereas the second distribution contains those induced by all other classes. In $k$-way learning problems, we estimated $p$-values for all $k$ classes and determined the minimum $p$-value for individual feature maps. Finally, the feature maps were selected for the final classification when the minimum $p$-values are smaller than the predefined threshold value ($P^{\theta}$), which is determined by the maximum threshold value ($P_{max}$) and $R$ (Eq. \ref{eq2}).

\begin{equation}
P^{\theta}_l=P_{max}\frac{R_l}{R_{max}}
\end{equation}
where l and $R_{max}$ denote layer i.d. and maximum relevance scores.

In the experiments, we noted the strong bias in class preference. That is, many feature maps selectively respond to popular classes, but only a fraction of feature maps respond to non-popular classes. This bias can lead to poor training of FCL. To avoid this problem, we selected the same number of features maps for all classes. Specifically, we estimated the numbers of selective feature maps for all classes and obtained the minimum number of feature maps ($N_{feature}$). Then, we limited the number of feature maps for all classes to be $N_{feature}$.

\subsection{Experimental protocol}
ATL selects feature maps and use their outputs to train FCL on target tasks with two hyperparameters, which are the maximal threshold for $p$-value $p_{max}$ and the number of the layers $N_{layer}$. We used $p_{max}=0.4$ and $N_{layer}=3$ in the experiments unless stated otherwise. For each benchmark dataset, we chose 30 classes and then used the first 5, 10, 15, 20, 25 and 30 classes to create 5, 10, 15, 20, 25 and 30-way learning problems. That is, a 5-way problem is a subset of 10-way or higher-way problems, and a 10-way problem is a subset of 15-way or higher-way problems. After selecting classes, 3, 5 and 10 examples were randomly drawn from all available training examples to train FCL, and the entire test set was used to evaluate FCL’s accuracy. In this study, we used the original test and training split when available. All models are implemented using Pytorch, open-source python DL library \cite{NEURIPS2019_9015}. FCL was trained with `Adam' optimizer \cite{adam_optim} for 50 epochs. The initial learning rate $\lambda$ was 0.01 and then was reduced by 20 \% at every 10 epochs; other parameters are adopted from the default values of Pytorch implementation of Adam optimizer. 

During training, we estimated FCL's accuracy on the test set at every 5 epoch and selected the best test accuracy as the FCL's accuracy. We trained 5 independent FCLs for ATL to calculate the average accuracy ($A_{ATL}$). In this study, to build the baseline model, we fine-tuned the last layer of ResNet50 on the same learning problems. For each problem, 5 baseline models were trained and tested to obtain the average accuracy ($A_{base}$). Finally, the performance gain of ATL ($G=A_{ATL}-A_{base}$) is estimated and reported. If ATL is more beneficial than the traditional FT, $A_{ATL}$ would be higher than the accuracy ($A_{base}$) of the baseline model. 

\section{Results}
This section describes the performance gain of ATL in the few-shot learning setting. In the experiments, we created 18 different learning problems (3-different shots $\times$  6 different-ways) from a single benchmark problem. Specifically, we tested 3-shot, 5-shot and 10-shot in 5, 10, 15, 20, 25 and 30 ways, and 5 benchmark problems were used.

\subsection{Relevance of ResNet50’s layers}
\begin{figure*}[t]
\centering
\includegraphics[width=1\textwidth]{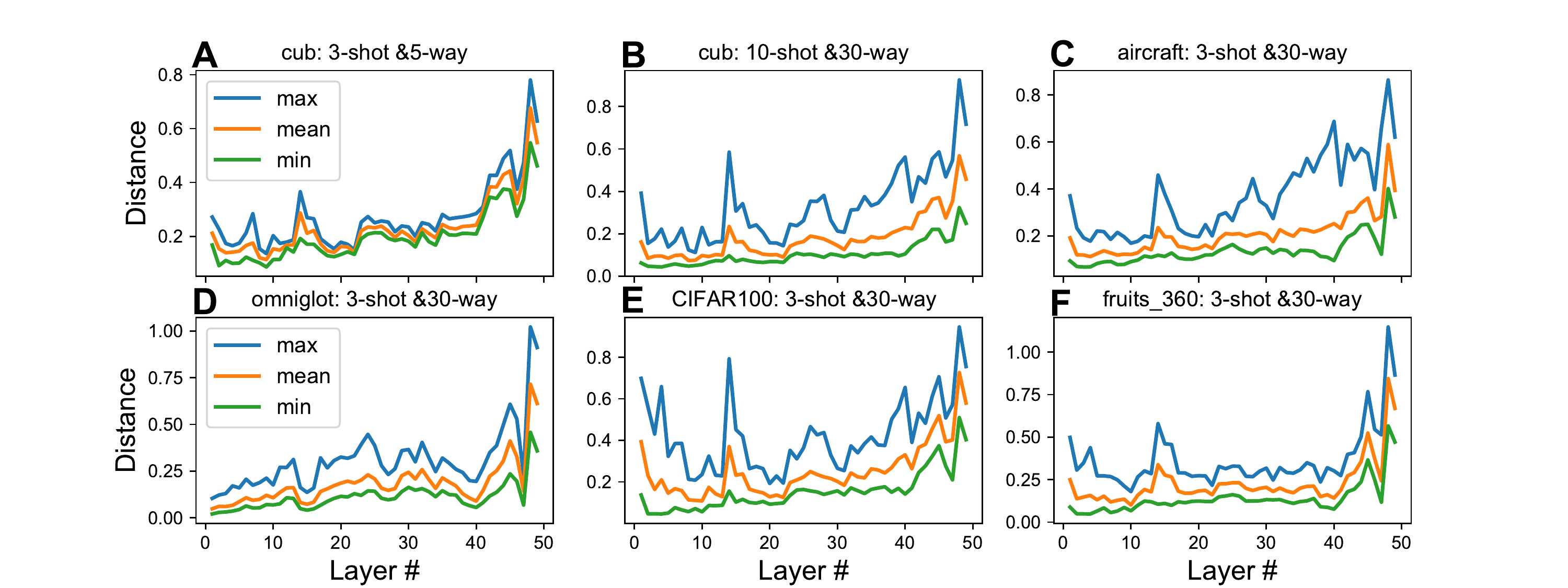}
\caption{Examples of layers’ relevance. ATL first evaluates the relevance of the layers to the target tasks. We display the estimated distance between classes' centroids. (A) and (B), Distance in two problems derived from cub. (B), (C), (D), (E) and (F), Distance in problems derived from aircraft, Omniglot, CIFAR100 and Fruits\_360, respectively. In the figure, maximum (blue), mean (red) and minimum (green) relevance scores are shown, but the minimum scores are used in the experiments. }
\label{fig2} 

\end{figure*}

To evaluate the relevance of ResNet50’s layers, we estimated the centroids of each class using the normalized activations ($LAV^r$) in individual layers (Eq. \ref{eq1}). Fig. \ref{fig2}A shows the maximum, mean and minimum distances in 48 layers in ResNet50, when 3 training examples are drawn from 5 classes of cub (CUB\_200\_2011); we excluded the first convolutional layer in our analysis. As shown in the figure, maximum, average and minimum distances show similar envelopes across layers, and the maximum distances vary the most strongly. As penultimate layers’ activations can reflect inputs’ classes \cite{snell_prototypical_2017,abdelkader_headless_2020}, we expected them to show the biggest separation (i.e., biggest distance) between classes, but the layer before the penultimate layer shows the greatest separation between classes. We further noted that 1) distances between centroids increase and decrease (roughly) periodically, as the layers go deeper and 2) top layers show homogeneous behaviors across problems (Fig. \ref{fig2}). In fact, the envelopes of the distances between the centroids are strikingly similar even across datasets (Fig. \ref{fig2}B-E). The reasons behind these observations remain unclear, but the results may still provide some insights into ResNet50’ learning process.

\subsection{Accuracy of ATL on cub}
\begin{figure*}[t]
\centering
\includegraphics[width=1\textwidth]{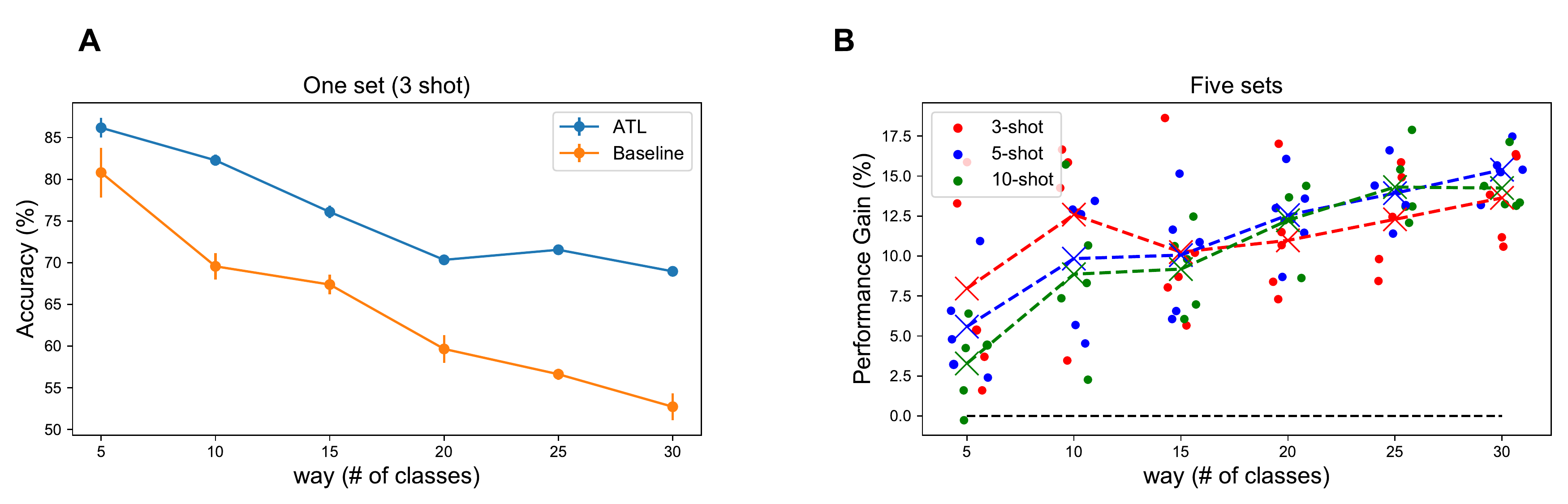}
\caption{Evaluation of ATL depending on the number of classes (i.e., way). (A), Accuracies of the baseline models and ATL on 3-shot learning problems derived the cub dataset. The mean values and standard deviations of 5 experiments are estimated with a single set. (B), Performance gain in all 5 sets of cub. Red, blue and green denotes 3-shot, 5-shot and 10-shot problems. The `x’ marks correspond to the mean values over all five sets. }
\label{fig3} 

\end{figure*}

By using relevance scores ($R$) and t-test (Sections 2.2 and 2.3),  we selected feature maps relevant to a 3-shot and 5-way learning problem derived from the cub dataset and used them to train ATL’s FCL (Fig. 1B). The mean values and standard deviations of accuracies of 5 ATL and the baseline models are displayed in Fig. \ref{fig3}A, suggesting that ATL outperforms the baseline model. We added additional classes to the training set until the way (number of classes to learn) reached 30. As shown in Fig. \ref{fig3}A, both ATL and the baseline model become less accurate, as the way increases. Importantly, ATL’s accuracy declines more slowly, and ATL consistently outperforms the baseline model. Further, the performance gain increases, when the way increases. 

The result above is encouraging, but we cannot rule out the possibility that the selected training set (30 classes from the cub dataset) selected could accidentally favor ATL over the baseline. Thus, we tested 4 more sets of classes and repeated the same experiments. It should be noted that the same parameters are used in all experiments. In each set, we estimated the performance gains ($G=A_{ATL}-A_{base}$) between ATL and the baseline model in all 90 learning problems (5 sets $\times$ 3 possible shots $\times $ 6 possible ways ). As shown in Fig. \ref{fig3}B, the performance gains in all 5 sets of cub dataset are positive except one (out of 90 problems). We made two germane observations. First, the performance gain and the way are also positively correlated, suggesting that ATL’s benefit grows when the way is bigger. Second, the performance gain is most pronounced in 3-shot learning problems. These results suggest that ATL can become more efficient when problems are more challenging.  

\subsection{Accuracy of ATL on other datasets}
Encouraged by the results, we applied ATL to learning problems derived from 4 more datasets (aircraft, CIFAR100, Fruits\_360 and Omniglot). DL’s performance varies significantly depending on hyperparameters, making empirical evaluation of DL difficult. Both ATL and the baseline models have multiple hyperparameters, which can be optimized for individual learning problems. In this study, instead of empirically optimizing them on each problem, we used the same parameters for all experiments, which are adopted from the cub experiments mentioned above. That is, the results described here may not be the optimal performances of both ATL and the baseline models, but we assume that a better learning algorithm (ATL or the baseline FT) can outperform, on average, with reasonable hyperparameters. Using 4 different datasets (360 learning problems derived from them) and the same hyperparameters, the comparison between ATL and the baseline should still be (approximately) accurate.

\begin{figure*}[t]
\centering
\includegraphics[width=1\textwidth]{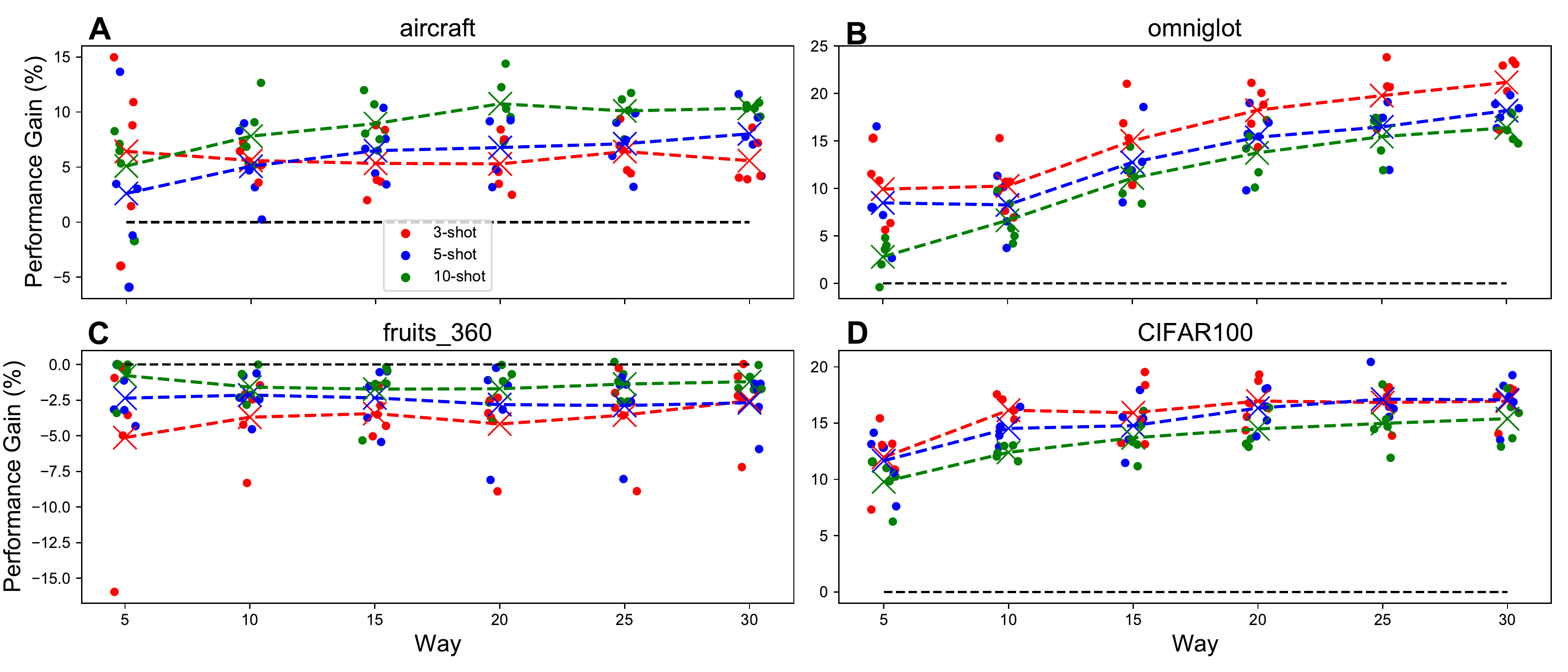}
\caption{Evaluation of ATL on 4 datasets. (A)-(D), Performance gain on problems derived from the aircraft, omniglot, fruits\_360 and CIFAR100. In all panels, red, blue and green denote 3-shot, 5-shot and 10-shot problems. The `x’ marks correspond to the mean values over all five sets. }
\label{fig4} 

\end{figure*}

\begin{figure*}[ht]
\centering
\includegraphics[width=1\linewidth]{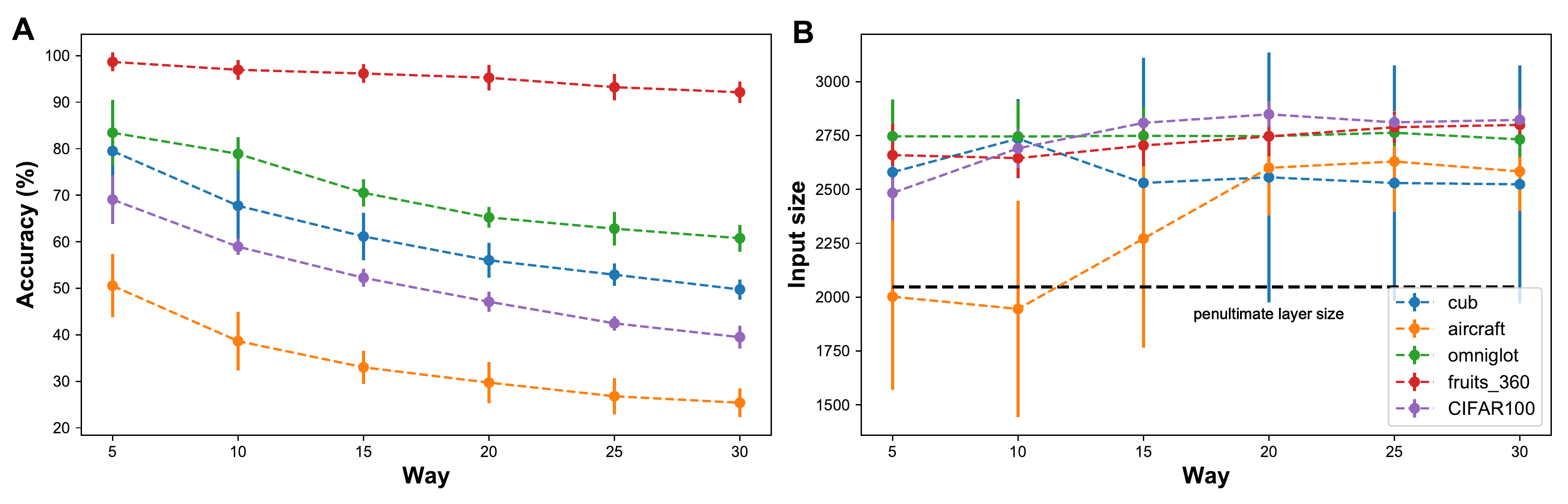}
\caption{(A), Accuracy of the baseline models on 3-shot problems derived from the cub (blue), aircraft (orange), omniglot (green), fruits\_360 (red) and CIFAR100 (purple). (B), Number of  selected feature maps for 3-shot learning problems. These numbers correspond to the size of input layers for FCL in ATL. The black dashed line represents the size of ResNet50’s penultimate layer (i.e, the input layer size for conventional FT’s FCL). }
\label{fig5} 

\end{figure*}

Fig. \ref{fig4} shows the performance gains in all 360 learning problems depending on the way and shot. Each dot represents the performance gain ($A_{ATL}-A_{base}$). We note that 1) ATL consistently outperforms the baseline models in learning problems derived from the aircraft dataset (FGVC-Aircraft), Omniglot and CIFAR 100 and 2) the performance gains on the problems derived from Fruits\_360 are negative. We further note that the performance gain is low, when the baseline model is highly accurate, but it is high when the baseline model becomes less accurate (Fig. \ref{fig5}A). For instance, when the baseline models are highly accurate (98.7\% for 5-way and 92.1 \% for 30-way problems) on learning problems derived from Fruits\_360 (see red line in Fig. \ref{fig5}A), ATL cannot provide any benefit (Fig. \ref{fig4}). The performance gain is higher when ATL is trained on learning problems derived from CIFAR100 (purple line in Fig. \ref{fig5}A). These results suggest that ATL can be an effective alternative TL solution when the conventional TL struggles in the few-shot learning setting. 

If feature maps are dynamically selected from multiple layers, a large number of feature maps could be chosen. In this case, it is necessary to train a large-scale FCL, which may reduce ATL’s utility. To assess ATL's required computing resources, we estimated the number of feature maps (i.e., the input size of ATL’s FCL) during 3-shot learning problems. As shown in Fig. \ref{fig5}B, the input size is largely comparable to the size of ResNet50’ penultimate layer. We observed that the input size does not increase rapidly even when the $N_{layer}$ and $P_{max}$ are higher than the default values. 

\subsection{ATL depending on hyperparameters}
\begin{figure*}[ht]
\centering
\includegraphics[width=1\textwidth]{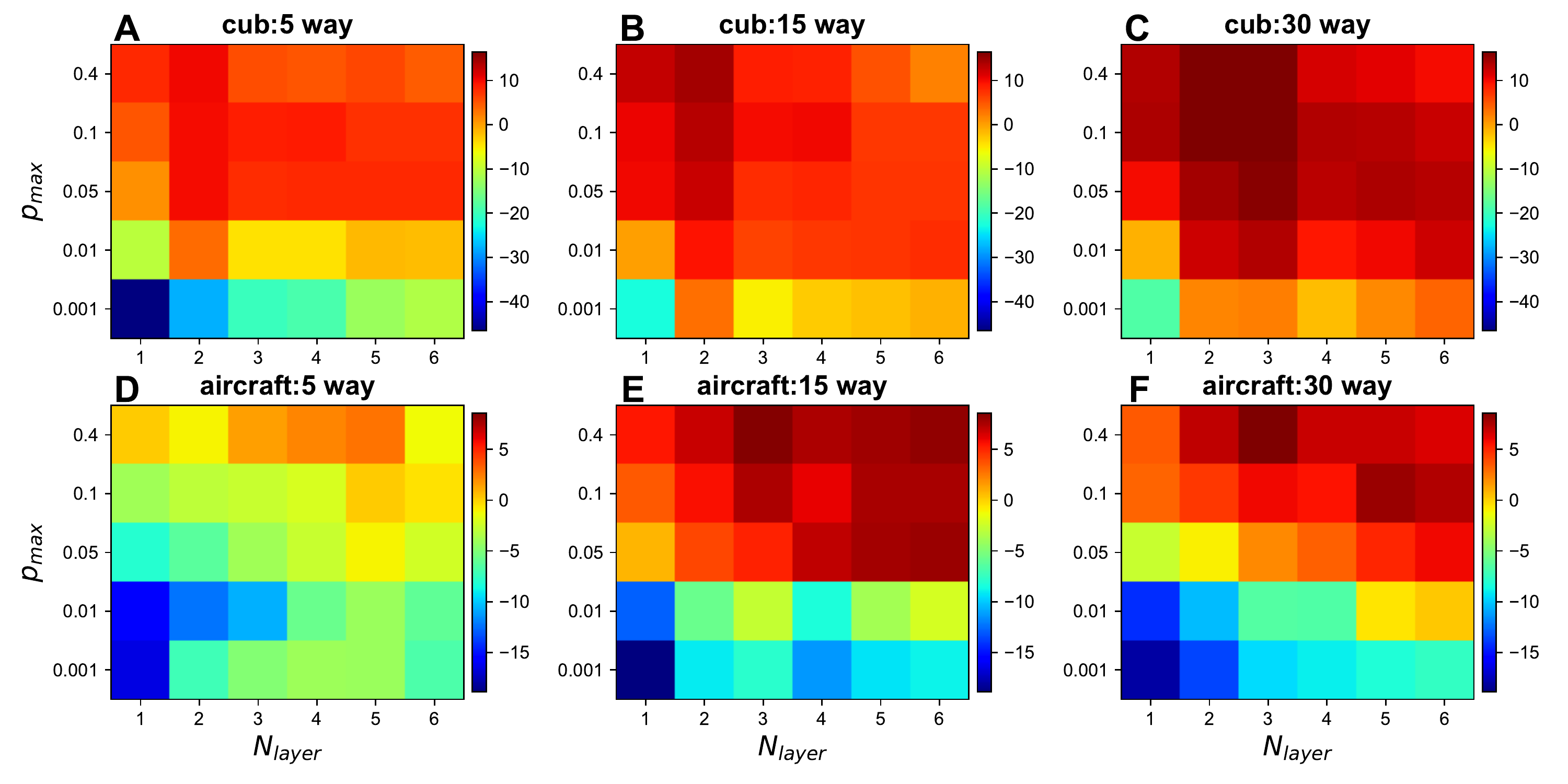}
\caption{Performance gain (\%) depending on the hyperparameters ($P_{max}$ and $N_{layer}$). X-axis and y-axis represent $N_{layer}$ and $P_{max}$. The performance gain is displayed using the color codes (see the color bars). (A)-(C), Performance gain on 5, 15 and 30 way problems derived from the cub dataset. (D)-(E), Performance gain on 5, 15 and 30 way problems derived from the aircraft dataset. }
\label{fig6} 

\end{figure*}

ATL requires two hyperparameters ($p_{max}$ and $N_{layer}$) to choose feature maps, and thus we evaluated the performance gain’s dependency on $p_{max}$ and $N_{layer}$ by using 3-shot learning problems derived from the cub and aircraft datasets. Fig. \ref{fig6} shows the performance gain depending on $N_{layer}$ and $p_{max}$. As shown in the figure, ATL outperforms the baseline model in a wide range of hyperparameters, suggesting that ATL does not require extensive fine-tuning of hyperparameters.

\subsection{Multilayer perceptrons as FCL}
\begin{figure*}[t]
\centering
\includegraphics[width=1\linewidth]{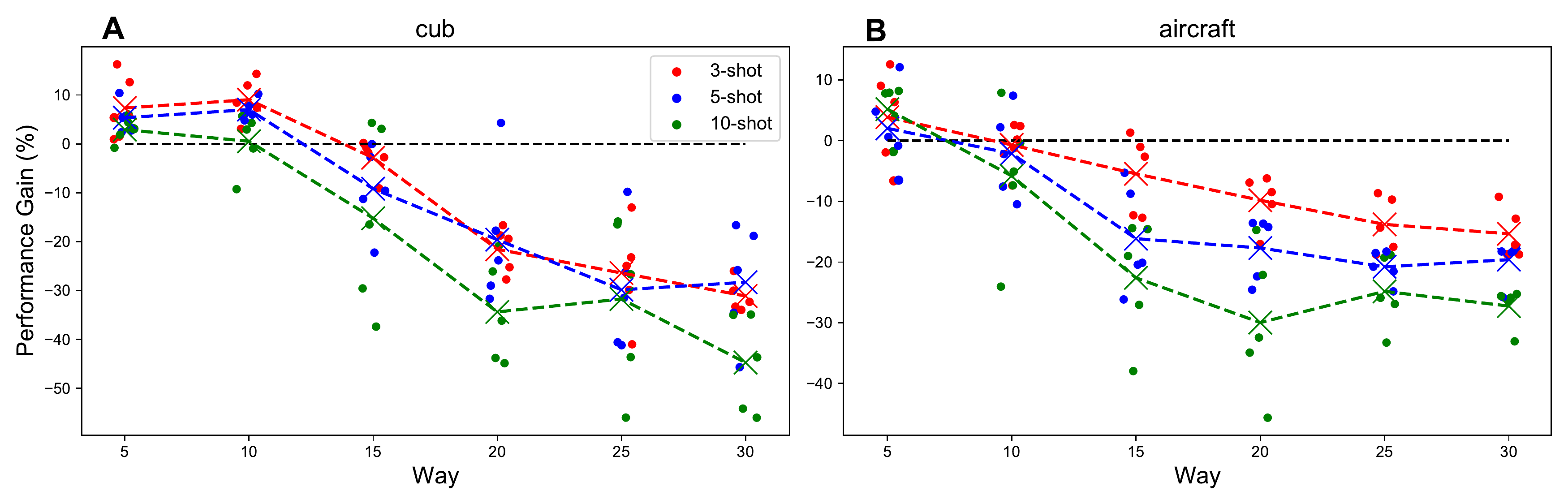}
\caption{Performance gain of ATL with multilayer perceptron as FCL. (A) and (B), Performance gain on the cub and aircraft datasets. }
\label{fig7} 

\end{figure*}

So far, we have used linear classifiers as FCL, but multilayer perceptrons can be used instead of linear classifiers. Thus, we evaluated ATL by replacing linear classifiers with single hidden layer perceptrons (MLP). The number of hidden neuron is 100, and their activation functions is the ReLU functions. Interestingly, the performance gain on aircraft and cub becomes negative, when the way becomes higher than 10 (Fig. \ref{fig7}). We note that ATL is 100\% accurate on training examples, when MLP is used, suggesting that MLP overfits to the outputs of the feature maps selected for ATL. That is, for ATL, the linear classifiers would be more suitable option than MLP.

\section{Discussion}
Here we asked if feature maps in earlier layers of pretrained teacher models can be used to train final classifiers on new target tasks more accurately. To address this question, we developed ATL that selects feature maps from relevant layers. Our empirical evaluations suggest that ATL can outperform the baseline model (the most common TL model) in few-shot learning problems (Figs. \ref{fig3} and \ref{fig4}) and that the input size of ATL’s FCL are comparable to the size of ResNet50’s penultimate layer (i.e., the input size of the baseline model’s FCL) (Fig. \ref{fig5}B). Based on these results, we propose that ATL can be an effective alternative to the traditional FT/TL, especially when only a few examples are available for a large number of classes (i.e., low-shot but high-way problems).

\subsection{ResNet’s operating principles}
The normalized distances in ResNet50’s layers are remarkably similar across datasets/problems (Fig. \ref{fig2}). Even top-layers of Omniglot, which is substantially different from other datasets, are almost identical to the top layers of other datasets, leading us to speculate that that the normalized distance may reflect the computations occurring in ResNet, instead of the properties of datasets. Specifically, we assume that ResNet repeatedly stretches and shrinks embedding spaces to find the best embeddings in penultimate layers to solve given problems, which we will address in the future. 

\subsection{Extension to other teacher models}
In this study, we tested ResNet50 as a possible teacher model because the earlier study \cite{veit_residual_nodate} suggested that ResNet can be considered as a collection of shallow models. However, it would be interesting to see if ATL can be extended to other popular pretrained models (teacher models) such as VGG19 \cite{simonyan_very_2015} or DenseNet \cite{huang_densely_2018}. If ATL can be applied to diverse pretrained models, we could extend ATL to aggregate diverse feature maps from multiple teacher models with different architectures to obtain higher performance gains. We plan to address this possibility in the future. 

\subsection{Limitation}
Lastly, we discuss the limitations of our study. First, we focused on comparing ATL and the baseline model under the same condition and did not perform full hyperparameter search for both ATL and the baseline models. That is, the performance discussed above may not reflect the truly optimal performances of both ATL and the baseline model. Second, ATL is not always better than the baseline model. In the experiments with Fruts\_360 problems, ATL is inferior to the baseline model that is highly accurate (Fig. \ref{fig5}A). Third, ATL needs to estimate the relevance of layers and feature maps, requiring additional overhead cost. The price of overhead cost may not be high (Fig. \ref{fig5}B), when ATL is applied to few-shot learning problems, but it would pose some challenges, when ATL is applied to problems with substantial amounts of data. Despite these limitations, our experiments raise the possibility that ATL can be a good alternative to the conventional TL, when the available data is limited.

%\section*{Acknowledgements}
%This work was funded by the U.S. Government.

\bibliographystyle{unsrt}
\bibliography{refs}{}

\begin{thebibliography}{10}

\bibitem{lecun_deep_2015}
Yann LeCun, Yoshua Bengio, and Geoffrey Hinton.
\newblock Deep learning.
\newblock {\em Nature}, 521(7553):436--444, May 2015.

\bibitem{lecun_gradient-based_1998}
Y~Lecun, L~Bottou, Y~Bengio, and P~Haffner.
\newblock Gradient-based learning applied to document recognition.
\newblock {\em PROCEEDINGS OF THE IEEE}, 86(11):47, 1998.

\bibitem{hertz_introduction_nodate}
John Hertz, Anders Krogh, and Richard Palmer.
\newblock {\em Introduction {To} {The} {Theory} {Of} {Neural} {Computation}}.
\newblock {SANTA} {FE} {INSTITUTE} {IIN} {THE} {SCIENCES} {OF} {COMPLEXITY}.
  CRC Press, 1991.

\bibitem{pan_survey_2010}
Sinno~Jialin Pan and Qiang Yang.
\newblock A {Survey} on {Transfer} {Learning}.
\newblock {\em IEEE TRANSACTIONS ON KNOWLEDGE AND DATA ENGINEERING}, 22(10):15,
  2010.

\bibitem{yosinski_how_nodate}
Jason Yosinski, Jeff Clune, Yoshua Bengio, and Hod Lipson.
\newblock How transferable are features in deep neural networks?, 2014.

\bibitem{azizpour_generic_nodate}
Hossein Azizpour, Ali~Sharif Razavian, Josephine Sullivan, Atsuto Maki, and
  Stefan Carlsson.
\newblock From {Generic} to {Specific} {Deep} {Representations} for {Visual}
  {Recognition}.
\newblock In {\em CVPR}, page~10, 2015.

\bibitem{kornblith_better_nodate}
Simon Kornblith, Jonathon Shlens, and Quoc~V Le.
\newblock Do {Better} {ImageNet} {Models} {Transfer} {Better}?
\newblock {\em CVPR}, 2019.

\bibitem{tajbakhsh_convolutional_2016}
Nima Tajbakhsh, Jae~Y Shin, Suryakanth~R Gurudu, R~Todd Hurst, Christopher~B
  Kendall, Michael~B Gotway, and Jianming Liang.
\newblock Convolutional {Neural} {Networks} for {Medical} {Image} {Analysis}:
  {Full} {Training} or {Fine} {Tuning}?
\newblock {\em IEEE TRANSACTIONS ON MEDICAL IMAGING}, 35(5):14, 2016.

\bibitem{razavian_cnn_nodate}
Ali~Sharif Razavian, Hossein Azizpour, Josephine Sullivan, and Stefan Carlsson.
\newblock {CNN} {Features} {Off}-the-{Shelf}: {An} {Astounding} {Baseline} for
  {Recognition}.
\newblock {\em CVPR}, pages 806--813, 2014.

\bibitem{veit_residual_nodate}
Andreas Veit, Michael Wilber, and Serge Belongie.
\newblock Residual {Networks} {Behave} {Like} {Ensembles} of {Relatively}
  {Shallow} {Networks}.
\newblock In {\em NIPS}, 2016.

\bibitem{he_deep_2015}
Kaiming He, Xiangyu Zhang, Shaoqing Ren, and Jian Sun.
\newblock Deep {Residual} {Learning} for {Image} {Recognition}.
\newblock {\em arXiv:1512.03385 [cs]}, December 2015.
\newblock arXiv: 1512.03385.

\bibitem{WahCUB_200_2011}
C.~Wah, S.~Branson, P.~Welinder, P.~Perona, and S.~Belongie.
\newblock {The Caltech-UCSD Birds-200-2011 Dataset}.
\newblock Technical Report CNS-TR-2011-001, California Institute of Technology,
  2011.

\bibitem{maji13fine-grained}
Subhransu Maji, Esa Rahtu, Juho Kannala, Matthew Blaschko, and Andrea Vedaldi.
\newblock Fine-grained visual classification of aircraft, 2013.

\bibitem{krizhevsky_learning_nodate}
Alex Krizhevsky.
\newblock Learning {Multiple} {Layers} of {Features} from {Tiny} {Images}.
\newblock {\em Tech Report}, page~60, 2019.

\bibitem{article}
Horea Mureșan and Mihai Oltean.
\newblock Fruit recognition from images using deep learning.
\newblock {\em Acta Universitatis Sapientiae, Informatica}, 10:26--42, 06 2018.

\bibitem{Lake1332}
Brenden~M. Lake, Ruslan Salakhutdinov, and Joshua~B. Tenenbaum.
\newblock Human-level concept learning through probabilistic program induction.
\newblock {\em Science}, 350(6266):1332--1338, 2015.

\bibitem{simonyan_very_2015}
Karen Simonyan and Andrew Zisserman.
\newblock Very {Deep} {Convolutional} {Networks} for {Large}-{Scale} {Image}
  {Recognition}.
\newblock {\em arXiv:1409.1556 [cs]}, April 2015.
\newblock arXiv: 1409.1556.

\bibitem{pmlr-v32-donahue14}
Jeff Donahue, Yangqing Jia, Oriol Vinyals, Judy Hoffman, Ning Zhang, Eric
  Tzeng, and Trevor Darrell.
\newblock Decaf: A deep convolutional activation feature for generic visual
  recognition.
\newblock In Eric~P. Xing and Tony Jebara, editors, {\em Proceedings of the
  31st International Conference on Machine Learning}, volume~32 of {\em
  Proceedings of Machine Learning Research}, pages 647--655, Bejing, China,
  22--24 Jun 2014. PMLR.

\bibitem{10.1007/978-3-319-10590-1_53}
Matthew~D. Zeiler and Rob Fergus.
\newblock Visualizing and understanding convolutional networks.
\newblock In David Fleet, Tomas Pajdla, Bernt Schiele, and Tinne Tuytelaars,
  editors, {\em Computer Vision -- ECCV 2014}, pages 818--833, Cham, 2014.
  Springer International Publishing.

\bibitem{8099808}
Yin Cui, Feng Zhou, Jiang Wang, Xiao Liu, Yuanqing Lin, and Serge Belongie.
\newblock Kernel pooling for convolutional neural networks.
\newblock In {\em 2017 IEEE Conference on Computer Vision and Pattern
  Recognition (CVPR)}, pages 3049--3058, 2017.

\bibitem{song_locally-transferred_2017}
Yang Song, Fan Zhang, Qing Li, Heng Huang, Lauren~J. O'Donnell, and Weidong
  Cai.
\newblock Locally-{Transferred} {Fisher} {Vectors} for {Texture}
  {Classification}.
\newblock In {\em 2017 {IEEE} {International} {Conference} on {Computer}
  {Vision} ({ICCV})}, pages 4922--4930, Venice, October 2017. IEEE.

\bibitem{gao_compact_2016}
Yang Gao, Oscar Beijbom, Ning Zhang, and Trevor Darrell.
\newblock Compact {Bilinear} {Pooling}.
\newblock In {\em 2016 {IEEE} {Conference} on {Computer} {Vision} and {Pattern}
  {Recognition} ({CVPR})}, pages 317--326, Las Vegas, NV, USA, June 2016. IEEE.

\bibitem{cimpoi_deep_2015}
Mircea Cimpoi, Subhransu Maji, and Andrea Vedaldi.
\newblock Deep filter banks for texture recognition and segmentation.
\newblock In {\em 2015 {IEEE} {Conference} on {Computer} {Vision} and {Pattern}
  {Recognition} ({CVPR})}, pages 3828--3836, Boston, MA, USA, June 2015. IEEE.

\bibitem{7410527}
Tsung-Yu Lin, Aruni RoyChowdhury, and Subhransu Maji.
\newblock Bilinear cnn models for fine-grained visual recognition.
\newblock In {\em 2015 IEEE International Conference on Computer Vision
  (ICCV)}, pages 1449--1457, 2015.

\bibitem{guo_adafilter_2019}
Yunhui Guo, Yandong Li, Liqiang Wang, and Tajana Rosing.
\newblock {AdaFilter}: {Adaptive} {Filter} {Fine}-tuning for {Deep} {Transfer}
  {Learning}.
\newblock {\em arXiv:1911.09659 [cs]}, December 2019.
\newblock arXiv: 1911.09659.

\bibitem{guo_spottune_nodate}
Yunhui Guo, Honghui Shi, Abhishek Kumar, Kristen Grauman, Tajana Rosing, and
  Rogerio Feris.
\newblock {SpotTune}: {Transfer} {Learning} {Through} {Adaptive}
  {Fine}-{Tuning}.
\newblock {\em CVPR}, pages 2661--2671, 2019.

\bibitem{8575475}
Romain Mormont, Pierre Geurts, and Raphaël Marée.
\newblock Comparison of deep transfer learning strategies for digital
  pathology.
\newblock In {\em 2018 IEEE/CVF Conference on Computer Vision and Pattern
  Recognition Workshops (CVPRW)}, pages 2343--234309, 2018.

\bibitem{snell_prototypical_2017}
Jake Snell, Kevin Swersky, and Richard Zemel.
\newblock Prototypical {Networks} for {Few}-shot {Learning}.
\newblock In {\em {NeurIPS}}, page~11, 2017.

\bibitem{abdelkader_headless_2020}
Ahmed Abdelkader, Michael~J. Curry, Liam Fowl, Tom Goldstein, Avi
  Schwarzschild, Manli Shu, Christoph Studer, and Chen Zhu.
\newblock Headless {Horseman}: {Adversarial} {Attacks} on {Transfer} {Learning}
  {Models}.
\newblock {\em ICASSP 2020 - 2020 IEEE International Conference on Acoustics,
  Speech and Signal Processing (ICASSP)}, pages 3087--3091, May 2020.
\newblock arXiv: 2004.09007.

\bibitem{NEURIPS2019_9015}
Adam Paszke, Sam Gross, Francisco Massa, Adam Lerer, James Bradbury, Gregory
  Chanan, Trevor Killeen, Zeming Lin, Natalia Gimelshein, Luca Antiga, Alban
  Desmaison, Andreas Kopf, Edward Yang, Zachary DeVito, Martin Raison, Alykhan
  Tejani, Sasank Chilamkurthy, Benoit Steiner, Lu~Fang, Junjie Bai, and Soumith
  Chintala.
\newblock Pytorch: An imperative style, high-performance deep learning library.
\newblock In H.~Wallach, H.~Larochelle, A.~Beygelzimer, F.~d\textquotesingle
  Alch\'{e}-Buc, E.~Fox, and R.~Garnett, editors, {\em Advances in Neural
  Information Processing Systems 32}, pages 8024--8035. Curran Associates,
  Inc., 2019.

\bibitem{adam_optim}
Diederik~P. Kingma and Jimmy~Lei Ba.
\newblock Adam: {A} {Method} for {Stochastic} {Optimization}.
\newblock In {\em {ICLR}}, 2015.

\bibitem{huang_densely_2018}
Gao Huang, Zhuang Liu, Laurens van~der Maaten, and Kilian~Q. Weinberger.
\newblock Densely {Connected} {Convolutional} {Networks}.
\newblock {\em arXiv:1608.06993 [cs]}, January 2018.
\newblock arXiv: 1608.06993.

\end{thebibliography}


\begin{thebibliography}{10}

\bibitem{lecun_deep_2015}
Yann LeCun, Yoshua Bengio, and Geoffrey Hinton.
\newblock Deep learning.
\newblock {\em Nature}, 521(7553):436--444, May 2015.

\bibitem{goodfellow2015explaining}
Ian~J. Goodfellow, Jonathon Shlens, and Christian Szegedy.
\newblock Explaining and harnessing adversarial examples, 2015.

\bibitem{wong2020rfml}
Lauren~J. Wong, William H. Clark~IV au2, Bryse Flowers, R.~Michael Buehrer,
  Alan~J. Michaels, and William~C. Headley.
\newblock The rfml ecosystem: A look at the unique challenges of applying deep
  learning to radio frequency applications, 2020.

\bibitem{oshea_over--air_2018}
Timothy~James O'Shea, Tamoghna Roy, and T.~Charles Clancy.
\newblock Over-the-{Air} {Deep} {Learning} {Based} {Radio} {Signal}
  {Classification}.
\newblock {\em IEEE Journal of Selected Topics in Signal Processing},
  12(1):168--179, February 2018.

\bibitem{8454504}
Fan Meng, Peng Chen, Lenan Wu, and Xianbin Wang.
\newblock Automatic modulation classification: A deep learning enabled
  approach.
\newblock {\em IEEE Transactions on Vehicular Technology}, 67(11):10760--10772,
  2018.

\bibitem{anderson2019deep}
Adam Anderson, Steven~R. Young, F.~Kyle Reed, and Jason~M. Vann.
\newblock Deep modulation (deepmod): A self-taught phy layer for resilient
  digital communications, 2019.

\bibitem{zhao2020adversarial}
Zhengyu Zhao, Zhuoran Liu, and Martha Larson.
\newblock Adversarial color enhancement: Generating unrestricted adversarial
  images by optimizing a color filter, 2020.

\bibitem{url:gnu-radio}
{GNU Radio Website}, {accessed 2021}.

\bibitem{ma_towards_2018}
Aleksander Ma.
\newblock {TOWARDS} {DEEP} {LEARNING} {MODELS} {RESISTANT} {TO} {ADVERSARIAL}
  {ATTACKS}.
\newblock page~23, 2018.

\bibitem{adesina_adversarial_2021}
Damilola Adesina, Chung-Chu Hsieh, Yalin~E. Sagduyu, and Lijun Qian.
\newblock Adversarial {Machine} {Learning} in {Wireless} {Communications} using
  {RF} {Data}: {A} {Review}.
\newblock {\em arXiv:2012.14392 [eess]}, August 2021.
\newblock arXiv: 2012.14392.

\end{thebibliography}

\end{document}